# WESSA at SemEval-2020 Task 9: Code-Mixed Sentiment Analysis using Transformers

**Ahmed Sultan**
WideBot
a.sultan@widebot.ai

**Mahmoud Salim**
WideBot
m.salim@widebot.ai

**Amina Gaber**
WideBot
amina.gaber@widebot.ai

**Islam El Hosary**
WideBot
islam@widebot.ai

## Abstract

In this paper, we describe our system submitted for SemEval 2020 Task 9, Sentiment Analysis for Code-Mixed Social Media Text alongside other experiments. Our best performing system is a Transfer Learning-based model that fine-tunes XLM-RoBERTa, a transformer-based multilingual masked language model, on monolingual English and Spanish data and Spanish-English code-mixed data. Our system outperforms the official task baseline by achieving a 70.1% average F1-Score on the official leaderboard using the test set. For later submissions, our system manages to achieve a 75.9% average F1-Score on the test set using CodaLab username "ahmed0sultan".

## 1 Introduction

Microblogging websites have been a huge source of data containing different kinds of information. Since the users on these microblogging websites tend to write informal real-time messages, they also tend to mix languages as they are just being spontaneous and want to ease the communication or they are just multilingual or non-native language speakers who mix between their native language and the language they are trying to use Patwa et al. (2020). This type of writing is called Code-Mixing or Code-Switching and it could be defined as the phenomenon of mixing the vocabulary and syntax of multiple languages in the same sentence Lal et al. (2019). Sentiment Analysis (SA) is the task of detecting, extracting, and classifying sentiment and opinions Montoyo et al. (2012). SA can help in structuring data of public opinions about products, brands or any topic that people can express opinions about, to be used in a very wide set of practical applications varying from political use, e.g. monitoring public events Tumasjan et al. (2010) to commercial use, e.g. making decisions in the stock market Jansen et al. (2009). The task of monolingual sentiment analysis has been a well-studied topic in the literature over the past few decades. However, little attention has been directed to SA based on code-mixed data.

In SemEval-2020 Task 9: Sentiment Analysis for Code-Mixed Social Media Text Patwa et al. (2020), the organizers provide a dataset of Code-Mixed tweets with word-level language labels that we will explore in Section 2.1, and with the following sentiment labels: positive, negative, neutral. Given a code-mixed text, the task is to classify the overall sentiment of the input text to one of the three sentiment labels mentioned above. The official evaluation metric for this task is Average F1-Score. We report the experiments made only on Spanish-English (Spanglish) data, whereas SemEval-2020 Task 9 contains Hindi-English (Hinglish) data as well. The challenges of this shared task could be summarized as follows: a) The relatively small dataset provided makes it hard to train complex models b) the target classes distribution is imbalanced in the training data c) the characteristics of social media text pose difficulties such as out-of-vocabulary words and ungrammatical sentences due to the spontaneous and informal writing with lots of extra embedded information in a single sentence (e.g. hashtags, emojis, or repeated characters in a word) some of this embedded information can be utilized to make a better predictive model and others may hurt the model prediction badly.

We conducted several experiments to tackle this problem. We used a Linear SVM, Logistic Regression, and Multinomial Naive Bayes models with TF-IDF feature vectors as an input. We also used XLM-RoBERTa Conneau et al. (2020), a transformer-based multilingual masked language model which is trained on 100 languages, fine-tuned on our downstream SA task which achieves our highest score outperforming the official baseline. The rest of the paper is structured as follows: Section 2 introduces some background about the task, an overview of the dataset, and related work. We describe the applied preprocessing steps and the experiments in Section 3. We report the results in Section 4. And Section 5 summarizes our work.

## 2 Background

Whereas the task of Sentiment Analysis has been extensively explored in the past few decades, there has been less work on Code-Mixed Sentiment Analysis (CMSA). Given a sentence in the code-mixed language, the task is to predict its sentiment polarity (i.e. positive, negative, or neutral). For example, the Spanglish tweet "ha u know is true parece ke te esta dando un atake D" should be classified as a positive sentiment.

### 2.1 Data

We use a combination of external monolingual English and Spanish data, and English-Spanish code-mixed (Spanglish) data which is provided by the shared task organizers, listed below and summarized in Table 1.

- **Spanglish Dataset**: The dataset provided by the shared task organizers Patwa et al. (2020) consists of SpanishEnglish code-mixed tweets with target labels for each tweet, namely, positive, negative, or neutral. Also, a word-level language annotation Molina et al. (2016) was provided with the following labels (*lang1, lang2, other, ne, unk, ambiguous, mixed, fw*), where *lang1* and *lang2* tags refer to English and Spanish language respectively. The *other* tag refers to emojis, emoticons, usernames, URLs, symbols, and punctuation marks. The *ne* tag is used to label named entities. The *unk* tag represents all the gibberish and unintelligible tokens. Words that can be either *lang1* or *lang2* are tagged as *ambiguous*. The *mixed* tag is used for code-mixed morphemes. Finally the *fw* tag represents words from a language that is neither *lang1* nor *lang2*. The dataset is divided into 12,002 labeled training samples, 2,998 labeled samples as development sets, and 3,789 unlabeled samples as a test set.
- **T4SA Dataset**: We use an external data resource to increase the training set and overcome the data imbalance in the shared task data which is a dataset of 12,000 English tweets from Twitter for Sentiment Analysis (T4SA) dataset Vadicamo et al. (2017) and we auto-translated them into 12,000 Spanish tweets using Yandex Translate API[1].

|          | Spanglish | | T4SA | | Total | |
|---|---|---|---|---|---|---|
|          | training | development | training | development | training | development |
| Negative | 2023 | 506 | 8000 | NA | 10023 | 506 |
| Neutral  | 3974 | 994 | 8000 | NA | 11974 | 994 |
| Positive | 6005 | 1498 | 8000 | NA | 14005 | 1498 |

Table 1: Datasets Description

### 2.2 Related work

With the rise of social media data, Sentiment Analysis became a field of interest for many researchers. Read (2005) demonstrated that dependency in sentiment classification can take the form of a domain, topic, temporal and language style. To perform Sentiment Analysis of the short text Dos Santos and Gatti (2014) proposed a deep convolutional neural network that exploits information from character to sentence level. Agarwal et al. (2011) examined two types of models: tree kernel and feature-based models.

---
[1] https://tech.yandex.com/translate/

In recent years, due to the tendency of some social media users to mix two or more languages while expressing their opinions, a new code-mixed language has been generated. A lot of research has been performed on code-mixed languages, such as Hindi-English Joshi et al. (2016) and English-Spanish Solorio and Liu (2008).
Sentiment Analysis of code-mixed data received a lot of attention from the research community. Different algorithms trained on code-mixed data. Mandal and Das (2018) used supervised learning algorithms, they trained Naïve Bayes (NB) model on English features only, and then they trained Support Vector Machine (SVM) model on code-mixed data as well as English features. Patra et al. (2018) trained Support Vector Machine (SVM) model on the word and character level n-grams features. Several supervised classifiers have been tested by Mandal and Das (2018) such as Gaussian Naïve Bayes (GNB), Bernoulli Naïve Bayes (BNB) and Multinomial Naïve Bayes (MNB) from the Naïve Bayes (NB) family. In addition, they had tested Linear Models (LM): Linear Regression (LRC) and Stochastic Gradient Descent (SGDC).

Deep learning-based techniques are widely used in detecting the sentiment of code-mixed data. Konate and Du (2018) showed that deep learning models outperformed the classical algorithms as deep learning models introduce more non-linearity in feature space, they tested with different models and their best model was one-layer Convolutional Neural Network (CNN). A sophisticated method utilizes the shared parameters of Siamese networks to map the sentences of code-mixed and standard languages to a common sentiment space that has been tested by Choudhary et al. (2018). A sub-word level LSTM (Subword-LSTM) architecture for learning sentiments in code-mixed data has been utilized in Joshi et al. (2016). Another line of research, Ghosh et al. (2017) they created their own dataset by manually labeling Facebook posts with their associated sentiments. They used word-based, semantic, and style-based features for classification. A Multilayer Perceptron model has been used to determine the polarity of the sentiment.

## 3 System Description

In this section, we are proposing different approaches to tackle the problem of identifying the sentiment in code-mixed text.

### 3.1 Preprocessing

As social media text is different from the regular text with irregular grammar, elongation, and all the informal usages of language, we needed to improve the quality of the text as it has a major effect on the final score. So we perform our preprocessing steps as follows:

- Remove emoticons and emojis and replace them with their textual meaning.

    "I love you so much , <3" → "I love you so much smiley face heart"

- Remove Mentions and non-ASCII Characters.

- Replace URLs with <URL> "The article URL is www.example.com" → "The article URL is URL"

- Remove elongation. "Hiiiii everyone" → "Hi everyone"

- Extract words from hashtags. "We need to talk #HereWeGoAgain" → "We need to talk Here We Go Again"

### 3.2 TF-IDF Vectorization

For the purpose of fitting the TF-IDF vectorizer, we used two different methods for preparing the input data:

- We treated each sentence in the datasets as a document so that we have input data size with the same size of the datasets mentioned in Section 2.1.

- For each sentiment label, we concatenated all the text so that we have an input of three rows representing the three sentiments (positive, neutral, negative).

For each one of the approaches mentioned above, we fitted two TF-IDF vectorizers on the whole text, the first vectorizer is a word-level TF-IDF vectorizer, and the second one is a character level TF-IDF vectorizer, then we concatenate the two vectorizers outputs into one vector which will be the input for the models discussed in Section 3.3 as illustrated in Figure1.

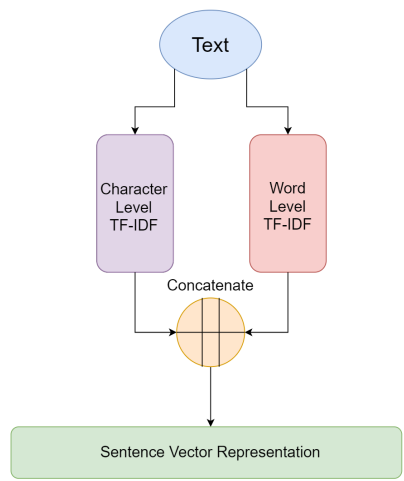

Figure 1: TF-IDF vectorization process

## 3.3 Machine Learning (ML) Models

We experimented with three different classical ML models: Logistic Regression (LR), Multinomial Naive Bayes (MNB), and Support Vector Machine (SVM) with a linear kernel Cortes and Vapnik (1995). Each model is trained on either the task dataset only or the T4SA data combined with the task dataset using the TF-IDF vectorization techniques explained in Section 3.2.

After several experiments with different settings, we found that increasing the train data size with the T4SA dataset doesn't have a noticeable increase in the evaluation metric perhaps that's because this dataset is from a different distribution which is monolingual, unlike the task dataset which is mostly code-mixed. Another reason could be the error propagation that might happen due to the automatic translation process, from English to Spanish, on the external dataset. So we only report the experiments conducted using the task dataset.

## 3.4 XLM-RoBERTa

Our best performing system is a fine-tuned XLM-RoBERTa Conneau et al. (2020), which is a general-purpose sentence representation and an extended version of mBERT and XLM (Lample and Conneau, 2019; Devlin et al., 2018) As they trained a transformer model Vaswani et al. (2017), with the multilingual Masked Language Model (MLM) objective using only monolingual data of 100 languages, on more data using Common-Crawl rather than using mere Wikipedia text. Using HuggingFace Transformers library Wolf et al. (2019), We fine-tuned XLM-RoBERTa base model on both Spanglish and T4SA training data in a randomly shuffled fashion. The preprocessing mentioned in Section 3.1 is used excluding 2 steps: the removal of non-ASCIIcharacters and word extraction from hashtags. This model is the submitted version which achieved a 70.1% average F1-Score on the official test set leaderboard.

Another version of the model described above, with slight changes in model's hyperparameters and preprocessing, achieved a 75.9% average F1-Score on the ofcial leaderboard test set. But due to the task limit number of submissions, this model wasn't considered on the official leaderboard.

We even enhance this model further after the submission phase by training the model on the monolingual data first then the code-mixed data Choudhury et al. (2017) achieving a higher score than its preceding versions on the development set rather than the test set since the test labels are never released.

# 4 Results

| System | TF-IDF Input | Dev Avg F1-Score |
|---|---|---|
| LR | concatenated docs per class | 51.60% |
| LR | all documents | 49.60% |
| MNB | concatenated docs per class | 50.40% |
| MNB | all documents | 50.73% |
| SVM | concatenated docs per class | 52.60% |
| SVM | all documents | 51.53% |
| **XLM-RoBERTa (submitted)** | NA | **52.20%** |
| **XLM-RoBERTa (enhanced)** | NA | **54.74%** |

Table 2: Evaluation results on development set

In this section, We report the results of our experiments in SemEval 2020 Task 9, Sentiment Analysis for Code-Mixed Social Media Text which introduces different systems, preprocessing steps, and data sources used. The official evaluation metric used is Average F1-Score. Due to the lack of test set labels, only the submitted system has a test set score, so we compare the different systems based on the development set scores. Table 2 shows the performance of all the systems described in the previous section. Compared with the other systems, the enhanced XLM-RoBERTa system achieves the highest score with a 54.74% average F1-Score on the development set outperforming the submitted XLM-RoBERTa system which achieved only 52.20%.

# 5 Conclusion

In this paper, we described our experimentations on SemEval 2020 Task 9, Sentiment Analysis for Code-Mixed Social Media Text. We presented four main systems; Logistic Regression, Multinomial Naive Bayes, SVM with linear kernel, and a fne-tuned XLM-RoBERTa. The submitted XLM-RoBERTa system outperformed the baseline on the test set with an average F1-Score of 70.1% and 52.20% on the development set. Whereas the enhanced approach of the XLMRoBERTa system outperformed the submitted version by achieving a 54.7% average F1-Score on the development set.

Future work includes working on using multilingual word embeddings that's aligned in the same space like MUSE Lample et al. (2018).